\documentclass[letterpaper,journal]{IEEEtran}
\usepackage{amsmath,amsfonts}
\usepackage{algorithmic}
\usepackage{algorithm}

\usepackage{array}
\usepackage[caption=false,font=normalsize,labelfont=sf,textfont=sf]{subfig}
\usepackage{textcomp}
\usepackage{newtxtext}
\usepackage{stfloats}
\usepackage{url}
\usepackage{verbatim}
\usepackage{graphicx}
\usepackage{tabularx}
\usepackage{tabularray}
\usepackage{booktabs}
\usepackage{cite}
\usepackage{hyperref}
\usepackage{xcolor,colortbl}
\definecolor{lightgray}{gray}{0.95}
\newcommand{\impr}[1]{\textcolor{green!60!black}{#1}}
\usepackage{multirow}
\usepackage{makecell}
\begin{document}

\title{High-Frequency First: A Two-Stage Approach for Improving\\Image INR}

\author{Sumit Kumar Dam, Mrityunjoy Gain, Eui-Nam Huh, and Choong Seon Hong,~\IEEEmembership{Fellow,~IEEE},
\thanks{Sumit Kumar Dam and Mrityunjoy Gain are with the Department of Artificial Intelligence, Kyung Hee University, Yongin-si 17104, Republic of Korea (email: skd160205@khu.ac.kr; gain@khu.ac.kr). Eui-Nam Huh and Choong Seon Hong are with the Department of Computer Science and Engineering, Kyung Hee University, Yongin-si 17104, Republic of Korea (email: johnhuh@khu.ac.kr; cshong@khu.ac.kr).\\
\indent Corresponding author: Choong Seon Hong.}
}



\maketitle
\begin{abstract}
Implicit Neural Representations (INRs) have emerged as a powerful alternative to traditional pixel-based formats by modeling images as continuous functions over spatial coordinates. A key challenge, however, lies in the spectral bias of neural networks, which tend to favor low-frequency components while struggling to capture high-frequency (HF) details such as sharp edges and fine textures. While prior approaches have addressed this limitation through architectural modifications or specialized activation functions, the potential of complementary training strategies remains relatively underexplored. In this work, we propose an orthogonal direction by directly guiding the training process. Specifically, we introduce a two-stage training strategy where a neighbor-aware soft mask adaptively assigns higher weights to pixels with strong local variations, encouraging early focus on fine details. The model then transitions to full-image training. Experimental results show that our approach is complementary to existing INR methods and consistently improves reconstruction quality. By assigning frequency-aware importance to pixels in image INR, our work offers an effective avenue to mitigate the spectral bias problem.
%\footnote{The source code is publicly available at \url{https://github.com/damnonymous/High-Frequency-First}}. 
\end{abstract}


\section{Introduction}
\IEEEPARstart{T}{he} field of signal representation has undergone a major transformation with the growing integration of deep learning techniques. Traditional methods typically rely on discretized formats, such as pixel matrices for images, point clouds for 3D scenes, or volumetric grids in simulations, to store and process signals \cite{genova2019learning,guo2020deep,wang2022survey}. While effective in many applications, these representations are constrained by resolution and often fail to reflect the continuous nature of real-world phenomena. In particular, the reliance on fixed-resolution sampling introduces limitations in scalability, memory efficiency, and reconstruction fidelity. Implicit Neural Representations (INRs) offer a more flexible alternative by modeling signals as continuous functions over spatial coordinates. Instead of storing discrete samples, INRs encode the signal in the parameters of a neural network, typically a multilayer perceptron (MLP), which learns a mapping from coordinates to signal values. This continuous representation enables resolution-agnostic reconstruction, supports adaptive sampling, and significantly lowers memory demands by eliminating the need for dense grids \cite{park2019deepsdf,chen2019learning,liu2020dist}. As a result, INRs have gained considerable attention across domains such as image representation, 3D modeling, and novel view synthesis, while emerging as a core paradigm in representation learning.

Although INR has shown great potential for continuous signal representation, one of its key limitations lies in the inherent tendency to prioritize smooth, low-frequency regions over finer details. This behavior, commonly known as spectral bias \cite{rahaman2019spectral}, causes inadequate learning of high-frequency information, which is often important for preserving boundaries and fine structures \cite{zheng2025exploring}. In image reconstruction tasks, this results in outputs that may appear overly smooth or lacking essential detail, even when trained on high-resolution inputs. Such frequency imbalance presents a serious challenge for tasks that require high spatial accuracy, and addressing it has become a key research direction for improving the overall fidelity of INR-based methods.

To address this issue, we propose a simple yet effective two-stage training strategy that improves INR performance by prioritizing high-frequency components in the early phase of learning. Specifically, we employ a neighbor-aware soft mask that adaptively assigns higher importance to pixels with strong local variations, such as those found in edges and textures. This selective emphasis guides the model to focus on complex structures during the initial stage of training, before transitioning to full-image supervision in the second stage. Experimental results demonstrate that our proposed strategy significantly improves PSNR and SSIM across multiple INR backbones on both natural images and CT scans. Our contributions are summarized as follows:
\begin{itemize}
    \item Unlike prior works that focus on modifying network architectures or activation functions, we introduce a two-stage training strategy, where a neighbor-aware soft mask encourages the model to emphasize high-frequency components before training on the full image.
    \item Our strategy is model-agnostic and seamlessly integrates with existing INR backbones.
    \item Experimental results on diverse image datasets demonstrate consistent improvements in PSNR and SSIM, validating the effectiveness of the proposed strategy.
\end{itemize}

Our code is available at \url{https://github.com/damnonymous/High-Frequency-First}. The rest of the paper is organized as follows: Section~\ref{sec:rel} reviews the related work. Section~\ref{sec:preli} introduces the preliminaries, including the formulation of INRs and evaluation metrics. Section~\ref{sec:method} presents the proposed methodology. Section~\ref{sec:exp} provides experimental results and analysis. Finally, Section~\ref{sec:concl} concludes the paper.

\section{Related Work}
\label{sec:rel}
\subsection{Implicit Neural Representations (INRs)}
INRs have emerged as a promising and increasingly popular framework for modeling signals in a continuous coordinate-based manner. Unlike traditional methods that discretize sampled data into pixels, volumes, or point clouds, INRs employ multilayer perceptrons (MLPs) to learn a mapping from spatial coordinates to signal values. This formulation enables compact, flexible, and resolution-agnostic representations, making INRs applicable across a wide range of domains, including image processing \cite{zheng2025exploring, chen2021learning}, 3D geometry \cite{genova2019learning,chen2019learning}, and generative modeling \cite{gu2021stylenerf,schwarz2020graf}.
%\vspace{-8mm}
\subsection{Spectral Bias and Existing Solutions}
Despite the advantages of INRs, their tendency to prioritize low-frequency components during training, known as spectral bias, remains a major challenge \cite{rahaman2019spectral,canatar2021spectral}. This bias leads to inadequate learning of high-frequency regions, resulting in reconstructions that often lack fine detail in tasks where structural fidelity is critical. Several techniques have been proposed to alleviate spectral bias. One prominent line of research focuses on modifying activation functions. For example, traditional INRs often employ ReLU due to its simplicity and empirical success; however, it lacks the capacity to effectively capture high-frequency signals. Sinusoidal Representation Networks (SIREN) \cite{sitzmann2020implicit} address this by introducing periodic sine-based activations, enabling better modeling of oscillatory patterns. Building on this idea, FINER \cite{liu2024finer} explores variable-periodic activation functions for flexible frequency tuning, while INCODE \cite{kazerouni2024incode} adopts harmonizer and composer networks to dynamically control sinusoidal activation parameters during training. WIRE \cite{saragadam2023wire} further improves signal representation by combining sinusoidal and Gaussian components through Gabor wavelet activations.
%Sinusoidal Representation Networks (SIREN) \cite{sitzmann2020implicit} address this by introducing periodic sine-based activations, enabling better modeling of oscillatory patterns. Building on this idea, FINER \cite{liu2024finer} explores dynamically tunable periodic functions, while INCODE \cite{kazerouni2024incode} adopts harmonizer and composer networks to control activation behavior during training. WIRE \cite{saragadam2023wire} further improves multiscale signal modeling by combining sinusoidal and Gaussian activations through Gabor wavelets.
Other approaches tackle spectral bias via input/output transformations, such as positional encodings with Fourier features \cite{tancik2020fourier}. Architectural modifications have also been explored to better capture signal components across spatial resolutions; these include multi-head decoders \cite{aftab2022multi}, partitioned sub-networks \cite{liu2023partition}, hierarchical structures such as BACON \cite{lindell2022bacon} and MINER \cite{saragadam2022miner}, as well as pyramid-based architectures like Pyramid NeRF \cite{zhu2023pyramid}. Beyond INR-specific solutions, researchers have also explored general optimization-oriented strategies, such as adaptive weighting and scheduling mechanisms. Prominent approaches include Focal Loss \cite{lin2017focal}, which emphasizes hard-to-classify samples based on prediction confidence; Uncertainty Weighting \cite{kendall2017uncertainties}, which guides model optimization using learned uncertainty estimates; and Curriculum Learning \cite{bengio2009curriculum}, which schedules training samples according to increasing difficulty. While these existing approaches primarily address spectral bias through activation designs, input encodings, architectural modifications, or optimization-focused strategies, our work specifically guides the training dynamics through a neighbor-aware soft mask that encourages the model to prioritize regions with stronger local variations before transitioning to full-image training.
% takes an orthogonal direction by our work
%In contrast to these approaches, our strategy takes an orthogonal direction by explicitly guiding the training dynamics, where we first emphasize the high-frequency components before transitioning to full-image training.

\section{Preliminary}
\label{sec:preli}
\subsection{Background}
Implicit Neural Representations (INRs) model signals as continuous functions by mapping spatial coordinates to corresponding output values through a neural network, typically a Multi-Layer Perceptron (MLP). Let $\mathbf{x} \in \mathbb{R}^d$ represent an input coordinate and $\mathbf{y} \in \mathbb{R}^c$ denote the corresponding signal value (e.g., RGB intensity). The INR is defined as a function $\psi_{\boldsymbol{\Theta}}: \mathbb{R}^d \rightarrow \mathbb{R}^c$, parameterized by $\boldsymbol{\Theta}$, such that:
\begin{equation}
\psi_{\boldsymbol{\Theta}}(\mathbf{x}) = \text{MLP}_{\boldsymbol{\Theta}}(\mathbf{x}).
\end{equation}

The MLP consists of a sequence of $L$ layers, where the transformation at the $\ell$-th layer is given by:
\begin{equation}
\mathbf{h}^{(\ell+1)} = \sigma\left( \mathbf{W}^{(\ell)} \mathbf{h}^{(\ell)} + \mathbf{b}^{(\ell)} \right).
\end{equation}
Here, $\mathbf{W}^{(\ell)}$ and $\mathbf{b}^{(\ell)}$ are the weight matrix and bias vector of the $\ell$-th layer, respectively, and $\sigma(\cdot)$ denotes a non-linear activation function, typically chosen from standard functions like ReLU or sine depending on the task.

Training the INR involves minimizing the difference between the predicted signal and the ground truth. This is typically achieved using the mean squared error (MSE) loss, defined as:
\begin{equation}
\mathcal{L}(\boldsymbol{\Theta}) = \frac{1}{N} \sum_{i=1}^{N} \left\| \psi_{\boldsymbol{\Theta}}(\mathbf{x}_i) - \mathbf{y}_i \right\|^2,
\end{equation}
where $\{ (\mathbf{x}_i, \mathbf{y}_i) \}_{i=1}^N$ is the set of input coordinates and their corresponding signal values. The parameters $\boldsymbol{\Theta}$ are updated via gradient descent:
\begin{equation}
\boldsymbol{\Theta} \leftarrow \boldsymbol{\Theta} - \eta \nabla_{\boldsymbol{\Theta}} \mathcal{L}(\boldsymbol{\Theta}),
\end{equation}
where $\eta$ represents the learning rate.

\begin{figure*}[htbp]
\centering
\includegraphics[width=0.905\linewidth]{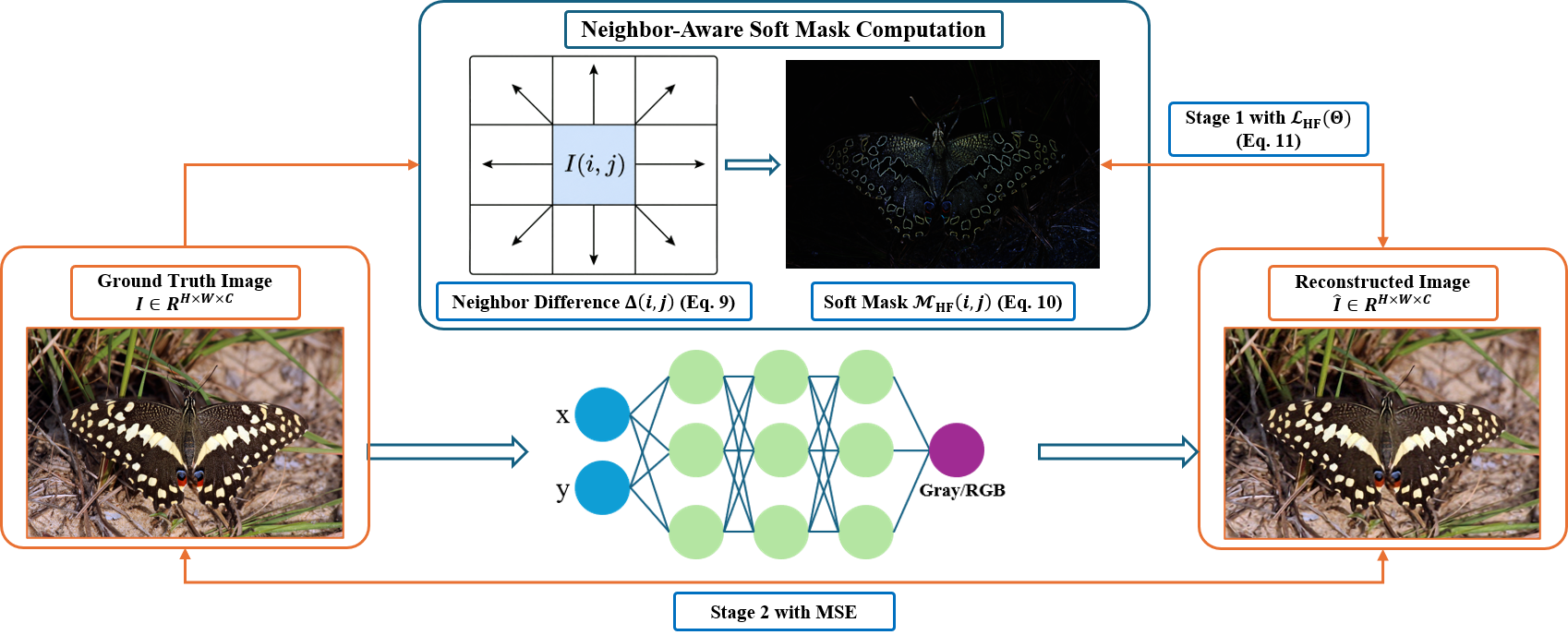}
\caption{Overview of our two-stage INR training strategy. The neighbor-aware soft mask is computed from local variations in the ground-truth image and used to prioritize high-frequency regions through $\mathcal{L}_{\mathrm{HF}}$ during Stage~1. Stage~2 continues training on the full image with standard MSE loss.}
\label{fig:framework}
\end{figure*}

\subsection{Evaluation Metrics}
To evaluate reconstruction quality, we adopt two widely used metrics: Peak Signal-to-Noise Ratio (PSNR) and Structural Similarity Index Measure (SSIM) \cite{wang2004image}. PSNR quantifies pixel-wise fidelity between the reconstructed and ground truth images. It is measured in decibels (dB) and is defined as:
\begin{equation}
\text{PSNR}(p, q) = 10 \cdot \log_{10} \left( \frac{MAX^2}{\text{MSE}(p, q)} \right),
\end{equation}
where $MAX$ is the maximum pixel value (e.g., 255 for 8-bit images), and MSE is the mean squared error computed as:
\begin{equation}
\text{MSE}(p, q) = \frac{1}{HW} \sum_{i=1}^{H} \sum_{j=1}^{W} (p_{ij} - q_{ij})^2.
\end{equation}

SSIM, on the other hand, measures perceptual similarity by comparing luminance, contrast, and structural components. It is defined as:
\begin{equation}
\text{SSIM}(p, q) = l(p, q) \cdot c(p, q) \cdot s(p, q),
\end{equation}
with each component expressed as:
\begin{equation}
\begin{aligned}
l(p, q) &= \frac{2\mu_p\mu_q + C_1}{\mu_p^2 + \mu_q^2 + C_1}, \\
c(p, q) &= \frac{2\sigma_p\sigma_q + C_2}{\sigma_p^2 + \sigma_q^2 + C_2}, \\
s(p, q) &= \frac{\sigma_{pq} + C_3}{\sigma_p \sigma_q + C_3}.
\end{aligned}
\end{equation}
Here, $\mu_p$ and $\mu_q$ are the mean intensities, $\sigma_p$ and $\sigma_q$ are the standard deviations, and $\sigma_{pq}$ is the covariance, all computed locally over a fixed window. The constants $C_1$, $C_2$, and $C_3$ are small positive values used to prevent numerical instability.

\section{Methodology}
\label{sec:method}
To address the spectral bias in INRs, we propose a two-stage training approach that initially guides the model to focus on high-frequency (HF) components of an image. The underlying assumption is that pixels exhibiting sharp local intensity variations, such as those found in edges, textures, or fine structures, carry high-frequency information. By identifying these pixels and prioritizing them early in training, we encourage the model to better capture fine details before transitioning to standard full-image optimization. This strategy gradually shifts the optimization emphasis across training stages, without modifying the network architecture. Fig.~\ref{fig:framework} provides an overview of our proposed method.

\subsection{Identifying High-frequency Pixels} Given an input image $I \in \mathbb{R}^{H \times W \times C}$, we apply symmetric padding to ensure valid comparisons across boundary pixels. Let $I_{\text{pad}}$ denote the padded image, and $I(i,j)$ denote the pixel at location $(i,j)$ in the original image. We compute the absolute difference between the center pixel and its neighbors within a fixed neighborhood $\mathcal{N}$, defined by a parameter $n$ that determines which neighbors are considered. For instance, $n = 4$ considers only horizontal and vertical neighbors, while $n = 8$ includes diagonal directions as well. The maximum absolute difference with neighbors at pixel $(i,j)$ is computed as:
\begin{equation}
\Delta(i,j) = \max_{(\Delta x,\Delta y)\in\mathcal{N}}
\left| I_{\text{pad}}(i+\Delta x,\,j+\Delta y) - I(i,j) \right|.
\end{equation}
%where $c \in \{R, G, B\}$ denotes the color channel.
This local difference captures sharp intensity variations, typically associated with fine structures and regions that are more challenging to fit during INR optimization.

\begin{algorithm}[!t]
\caption{Neighbor-Aware Soft Mask Computation}
\label{alg:softmask}
\begin{algorithmic}[1]
\REQUIRE ${I} \in \mathbb{R}^{H \times W \times C}$, $\tau$, $\alpha$, $n$
\ENSURE Soft mask $\mathcal{M}_{\text{HF}} \in [0,1]^{H \times W \times C}$
\STATE Pad the image symmetrically to obtain ${I}_{\text{pad}}$
\STATE Initialize $\Delta(i,j) \gets 0$
\STATE Define neighborhood $\mathcal{N}$ based on $n$
\FOR{each $(\Delta x, \Delta y) \in \mathcal{N}$}
    \STATE Compute: $d(i,j) \gets |{I}_{\text{pad}}(i+\Delta x, j+\Delta y) - {I}(i,j)|$
    \STATE Update: $\Delta(i,j) \gets \max(\Delta(i,j), d(i,j))$
\ENDFOR
\STATE Apply sigmoid: $\mathcal{M}_{\text{HF}}(i,j) \gets \sigma(\alpha \cdot (\Delta(i,j) - \tau))$
\RETURN $\mathcal{M}_{\text{HF}}$
\end{algorithmic}
\end{algorithm}

A soft mask $\mathcal{M}_{\text{HF}} \in [0,1]^{H \times W \times C}$ is used to assign a weight to each pixel based on its maximum absolute difference with neighboring pixels. The mask is computed as:
\begin{equation}
\mathcal{M}_{\text{HF}}(i,j) =
\sigma\!\left( \alpha \cdot \left( \Delta(i,j) - \tau \right) \right),
\end{equation}
where $\sigma(\cdot)$ is the sigmoid function, $\tau$ is a predefined threshold, and $\alpha$ controls the sharpness of the activation.

\indent During the early training stage, we apply this soft mask to emphasize high-frequency pixels. The loss is defined as:
\begin{equation}
\mathcal{L}_{\text{HF}}(\boldsymbol{\Theta}) =
\mathbb{E}_{i,j}
\left[
\mathcal{M}_{\text{HF}}(i,j)
\left\|
\psi_{\boldsymbol{\Theta}}(i,j) - I(i,j)
\right\|^2
\right].
\label{eq:hf_loss}
\end{equation}
Here, $\psi_{\boldsymbol{\Theta}}(i,j)$ denotes the model’s prediction at $(i,j)$ given parameters $\boldsymbol{\Theta}$. Algorithm \ref{alg:softmask} outlines the steps for neighbor-aware soft mask computation.\footnote{For multi-channel images, all operations are applied independently to each channel. We omit the channel index in the equations for simplicity.}

\subsection{Two-stage Training}
Following the computation of the neighbor-aware soft mask $\mathcal{M}_{\text{HF}}$, we adopt a two-stage training strategy. In the first stage, the model is trained using the loss $\mathcal{L}_{\text{HF}}$, where pixel-wise reconstruction errors are weighted by $\mathcal{M}_{\text{HF}}$ to prioritize high-frequency regions such as edges and textures. In the second stage, we switch to the standard mean squared error (MSE) loss over the full image, allowing the model to enhance the overall fidelity of the reconstruction. We use the same network architecture in both stages, with the training objective being the only component that changes. Further implementation details, including the number of training epochs for each stage, datasets, and model backbones, are provided in the next section.

\begin{table*}[t]
\centering
\caption{Quantitative results on \texttt{Kodak} and \texttt{Urban100} datasets. Best results are highlighted in bold. $\Delta$ indicates the performance gain of HF variants over their vanilla counterparts, reported as +PSNR (dB) / +SSIM.}
\label{tab:kodak_Urban100_flipped}
\resizebox{\textwidth}{!}{%
\begin{tabular}{c|cccccc|cccccc}
\toprule
\multirow{2}{*}{Metric} & \multicolumn{6}{c|}{KODAK} & \multicolumn{6}{c}{Urban100} \\
 & SIREN & \textbf{HF-SIREN} & FINER & \textbf{HF-FINER} & SCONE & \textbf{HF-SCONE} & SIREN & \textbf{HF-SIREN} & FINER & \textbf{HF-FINER} & SCONE & \textbf{HF-SCONE} \\
\midrule
PSNR & 33.51 & \textbf{36.78} & 37.07 & \textbf{40.22} & 33.18 & \textbf{40.66} & 23.17 & \textbf{24.46} & 24.82 & \textbf{26.47} & 21.61 & \textbf{23.59} \\
SSIM & 0.9185 & \textbf{0.9517} & 0.9589 & \textbf{0.9746} & 0.9403 & \textbf{0.9795} & 0.9005 & \textbf{0.9228} & 0.9274 & \textbf{0.9478} & 0.8640 & \textbf{0.9068} \\
$\Delta$ 
& -- & \impr{+3.27 / +0.0332}
& -- & \impr{+3.15 / +0.0157}
& -- & \impr{+7.48 / +0.0392}
& -- & \impr{+1.29 / +0.0223}
& -- & \impr{+1.65 / +0.0204}
& -- & \impr{+1.98 / +0.0428} \\
\bottomrule
\end{tabular}
}
\end{table*}

\begin{figure*}[t]
\centering
\includegraphics[width=.99\linewidth]{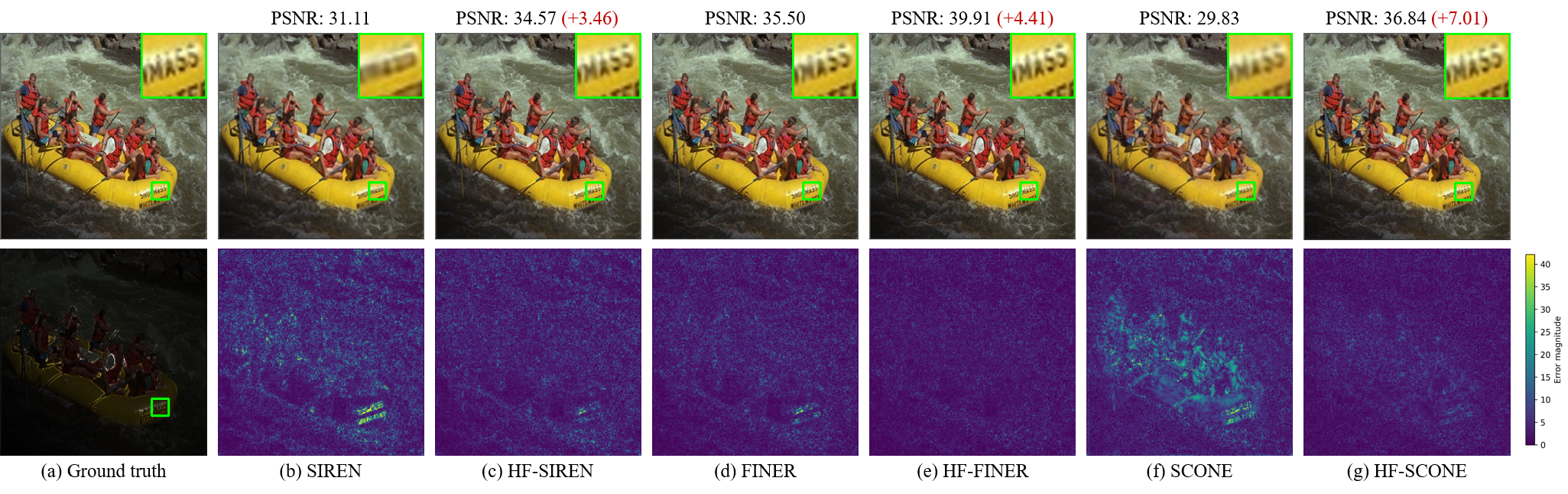}
\caption{Qualitative results on the \texttt{Kodak} dataset for different INR backbones and their HF variants (reference image: \texttt{kodim14}). The first column presents the ground-truth image (top) and the corresponding neighbor-aware HF soft mask extracted from the ground truth (bottom), with the green box indicating the selected HF text region (“MASS”) for zoomed-in comparison. The remaining columns show the reconstructed images (top) alongside the zoomed-in HF region and the corresponding error maps (bottom). PSNR values are reported for the reference image, and improvements of HF variants are shown in red.}
\label{fig:kodakq}
\vspace{-2.5mm}
\end{figure*}

\section{Experiments}
\label{sec:exp}
\subsection{Experimental Setup}
We conduct a series of experiments to evaluate the effectiveness of our two-stage training strategy across three vanilla INR backbones: SIREN~\cite{sitzmann2020implicit}, FINER~\cite{liu2024finer}, and SCONE~\cite{li2024learning}. These experiments include both natural images and CT scans. For clarity, we denote each baseline model combined with our module using the prefix “HF-”, resulting in HF-SIREN, HF-FINER, and HF-SCONE. To ensure consistent evaluation, every INR model is configured with three hidden layers, each comprising 256 neurons. All other parameters are set to the default configurations of the respective backbone implementations. Experiments are performed on a single NVIDIA RTX A5000 GPU using PyTorch 1.11.0. These models are trained using the Adam optimizer with a fixed learning rate of 1e-4. Unless otherwise specified, we train each model for 500 epochs. While baseline models use the standard MSE loss throughout training, HF-enhanced models follow a two-stage schedule: the first 250 epochs (Stage~1) utilize the proposed $\mathcal{L}_{\text{HF}}$ loss, followed by full-image training with the standard MSE loss for the remaining 250 epochs (Stage~2).

\subsection{Natural Image Reconstruction}
\begin{figure*}[t]
\centering
\includegraphics[width=.99\linewidth]{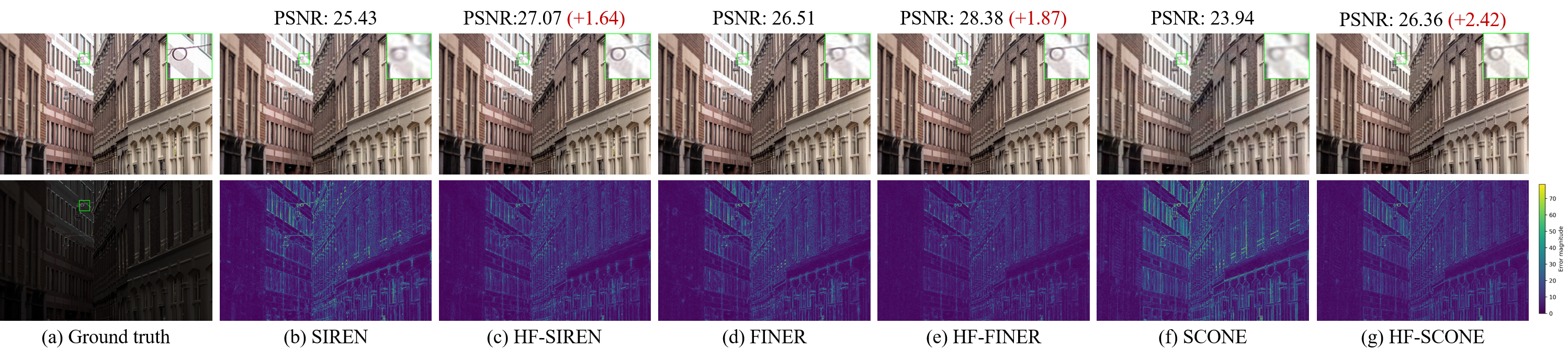}
\caption{Qualitative results on the \texttt{Urban100} dataset for different INR backbones and their HF variants (reference image: \texttt{img\_064\_SRF\_4\_HR}). Layout follows the format of Fig.~\ref{fig:kodakq}. The bottom-left image shows the neighbor-aware HF soft mask, with the green box indicating the target HF region (service loop) for zoomed-in comparison. Red values denote PSNR gains of HF variants over baselines.}
\label{fig:Urban100q}
\vspace{-1.25mm}
\end{figure*}

We first evaluate our method on the \texttt{Kodak} dataset~\cite{kodak_franzen}, a widely used benchmark consisting of 24 natural images. For this task, we resize the images from their original $768\times512$ resolution to $256\times256$. As summarized in Table~\ref{tab:kodak_Urban100_flipped}, the HF-enhanced variants consistently outperform their vanilla counterparts in terms of both PSNR and SSIM across all three backbones. In particular, HF-SCONE achieves a PSNR of $40.66$ dB, marking a substantial $+7.48$ dB improvement over its baseline, alongside a $+0.0392$ gain in SSIM. HF-SIREN and HF-FINER also exhibit notable improvements over their corresponding vanilla models. Qualitative results in Fig.~\ref{fig:kodakq} confirm these findings, where HF-enhanced models preserve the high-frequency text region (“MASS”) that appears noticeably blurred in the vanilla reconstructions. To further validate the scalability and generalization of our method, we conduct experiments on the higher-resolution \texttt{Urban100} dataset~\cite{huang2015single}. We select 20 images, each with a resolution of $680\times1024$, and perform the evaluation without resizing or modifying the training configurations. As shown in Table~\ref{tab:kodak_Urban100_flipped}, the HF-enhanced models retain their performance edge. Notably, HF-FINER reaches $26.47$ dB PSNR, outperforming the baseline FINER by $1.65$ dB, while HF-SCONE achieves the highest SSIM improvement ($+0.0428$). Visual comparisons in Fig.~\ref{fig:Urban100q} further highlight the effectiveness of our approach on complex urban scenes. Compared to baseline models, which lack textures and exhibit artifacts around fine architectural details such as window grids and building edges, our method better preserves these structures, yielding sharper and more faithful reconstructions.
\vspace{-1.5mm}

\begin{figure*}[t]
\centering
\includegraphics[width=.99\linewidth]{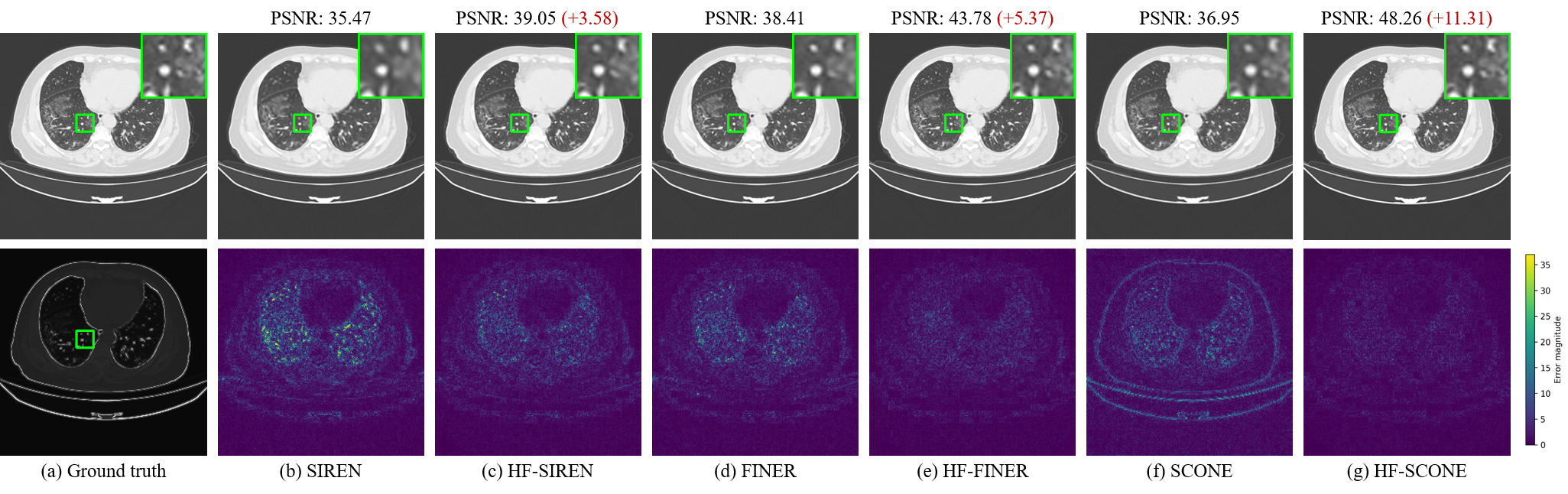}
\caption{Qualitative results on the \texttt{COVIDx CT-3A} dataset for different INR backbones and their HF variants (reference image: \texttt{coronacases\_001-0118}). Layout mirrors the format of Fig.~\ref{fig:kodakq}. The bottom-left image shows the neighbor-aware HF soft mask, with the green box indicating the target HF lung parenchymal region for zoomed-in comparison. Red values show PSNR gains over baselines.}
\label{fig:covidxctq}
\vspace{-2.5mm}
\end{figure*}

\begin{table}[t]
\centering
\caption{Quantitative results on \texttt{COVIDx CT-3A} images. Best results are highlighted in bold. $\Delta$ indicates the performance gain of HF variants over their baselines, reported as +PSNR (dB) / +SSIM.}
\label{tab:ct_column_delta}
\resizebox{.7\linewidth}{!}{%
\begin{tabular}{c|cc|c}
\toprule
\multirow{2}{*}{Model} & \multicolumn{2}{c|}{COVIDx CT-3A} & \multirow{2}{*}{$\Delta$} \\
& PSNR & SSIM & \\
\midrule
SIREN     & 35.73 & 0.9308 & -- \\
\textbf{HF-SIREN}  & \textbf{38.29} & \textbf{0.9479} & \impr{+2.56 / +0.0171} \\
\midrule
FINER     & 38.45 & 0.9557 & -- \\
\textbf{HF-FINER}  & \textbf{41.59} & \textbf{0.9707} & \impr{+3.14 / +0.0150} \\
\midrule
SCONE     & 37.64 & 0.9608 & -- \\
\textbf{HF-SCONE}  & \textbf{48.53} & \textbf{0.9941} & \impr{+10.89 / +0.0333} \\
\bottomrule
\end{tabular}
}
\vspace{-2.5mm}
\end{table}

\subsection{CT Image Reconstruction}
We now extend image reconstruction to the medical domain using the \texttt{COVIDx CT-3A} dataset~\cite{covidx_ct3a}. For the evaluation, we sample 20 images with an original resolution of $512\times512$, resized to $256\times256$. As reported in Table~\ref{tab:ct_column_delta}, the HF-enhanced models achieve superior reconstruction quality compared to their respective baselines. Specifically, HF-SCONE achieves a remarkable PSNR of $48.53$ dB, corresponding to a $+10.89$ dB improvement over vanilla SCONE, alongside an SSIM of $0.9941$. HF-SIREN and HF-FINER also show clear gains in both metrics. Fig.~\ref{fig:covidxctq} further illustrates that our training strategy better preserves complex structures, whereas baseline models tend to blur those regions.
\vspace{-1mm}

\subsection{Frequency Domain Analysis}
%Since the proposed method identifies high-frequency regions based on spatial local variations, we further validate its effectiveness in the frequency domain through Fourier analysis.
In this subsection, we perform frequency-domain analysis using the Fourier transform to evaluate the effectiveness of HF-INR in mitigating spectral bias.

\begin{figure*}[t]
\centering
\includegraphics[width=.835\linewidth]{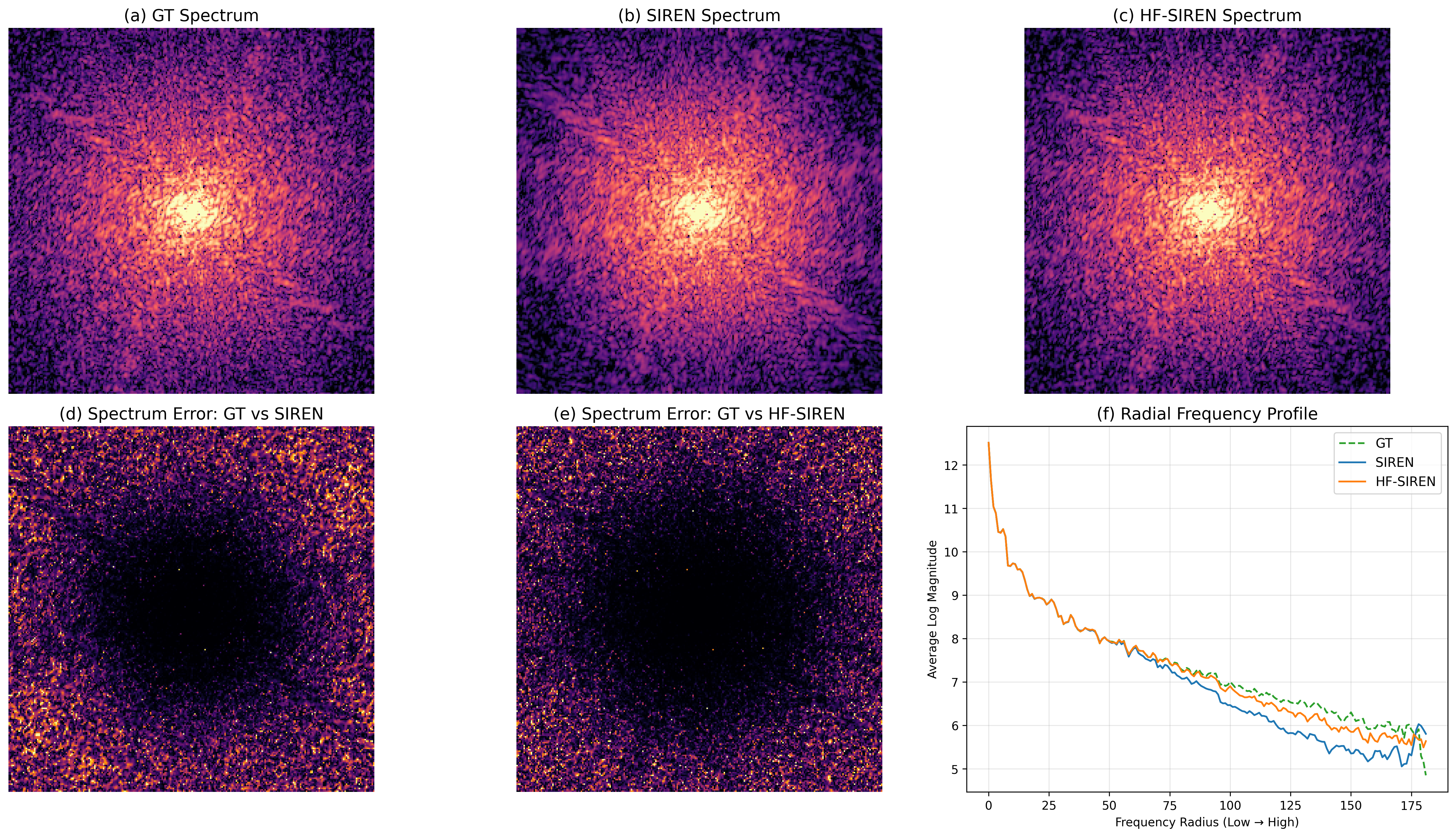}
\caption{Fourier spectrum analysis on \texttt{kodim14}. The top row shows the log-magnitude spectra of (a) GT, (b) SIREN, and (c) HF-SIREN. The bottom row presents spectrum error maps for (d) GT vs. SIREN and (e) GT vs. HF-SIREN, alongside (f) the radial frequency profile.}
%Fourier spectrum analysis on \texttt{kodim14}. The top row displays log-magnitude spectrum visualizations for (a) ground truth, (b) SIREN, and (c) HF-SIREN. The bottom row presents spectrum error maps for (d) GT vs. SIREN and (e) GT vs. HF-SIREN, alongside (f) the radial frequency profile.}
\label{fig:FFTAnalysis}
\vspace{-2.5mm}
\end{figure*}

\textbf{Fourier Spectrum Analysis.} Fig.~\ref{fig:FFTAnalysis} presents the frequency-domain comparison between SIREN and HF-SIREN on \texttt{kodim14}. The spectra are obtained by applying a 2D Fourier transform to the reconstructed images, followed by magnitude computation and logarithmic scaling. The log-magnitude spectra in (a)--(c) show that both methods recover the dominant low-frequency components, while noticeable differences emerge in the high-frequency regions. Compared with the ground truth, SIREN exhibits weaker responses in the outer spectral regions, whereas HF-SIREN preserves stronger high-frequency responses. This observation is further supported by the spectrum error maps in (d) and (e), which visualize the log-magnitude spectral differences with respect to the ground truth. Furthermore, for a quantitative comparison, we compute the radial frequency profile in Fig.~\ref{fig:FFTAnalysis}(f) by averaging the spectral magnitude over concentric frequency bands. While both methods align closely with the ground truth at low frequencies, SIREN exhibits a sharper decay at higher frequencies. In contrast, HF-SIREN maintains a closer correspondence to the ground truth across the high-frequency range, demonstrating the effectiveness of HF-INR in mitigating spectral bias.
%To demonstrate the effectiveness of HF-INR in mitigating spectral bias, we conduct frequency-domain analysis. Fig.~\ref{fig:FFTAnalysis} shows the spectral representation on \texttt{kodim14}. The spectra are obtained by applying a 2D Fourier transform to the reconstructed images, followed by magnitude computation and logarithmic scaling for visualization. The radial frequency profiles are computed by averaging the spectral magnitude over concentric frequency bands. The log-magnitude spectra in (a)–(c) indicate that all methods capture the dominant low-frequency components, while clear differences arise in the higher-frequency regions. Compared to the ground truth, the SIREN baseline exhibits attenuated responses in the outer spectral regions, reflecting limited recovery of high-frequency content. In contrast, HF-SIREN preserves stronger high-frequency components, suggesting improved reconstruction of fine details. This observation is further supported by the spectrum error maps in (d) and (e), which visualize the log-magnitude spectral differences with respect to the ground truth. The radial frequency profiles in (f) provide a quantitative comparison, showing that while both methods align closely with the ground truth at low frequencies, SIREN exhibits a sharper decay at higher frequencies. HF-SIREN maintains a closer match across the high-frequency range, confirming its effectiveness in mitigating spectral bias.

\begin{figure}[t]
\centering
\includegraphics[width=.945\linewidth]{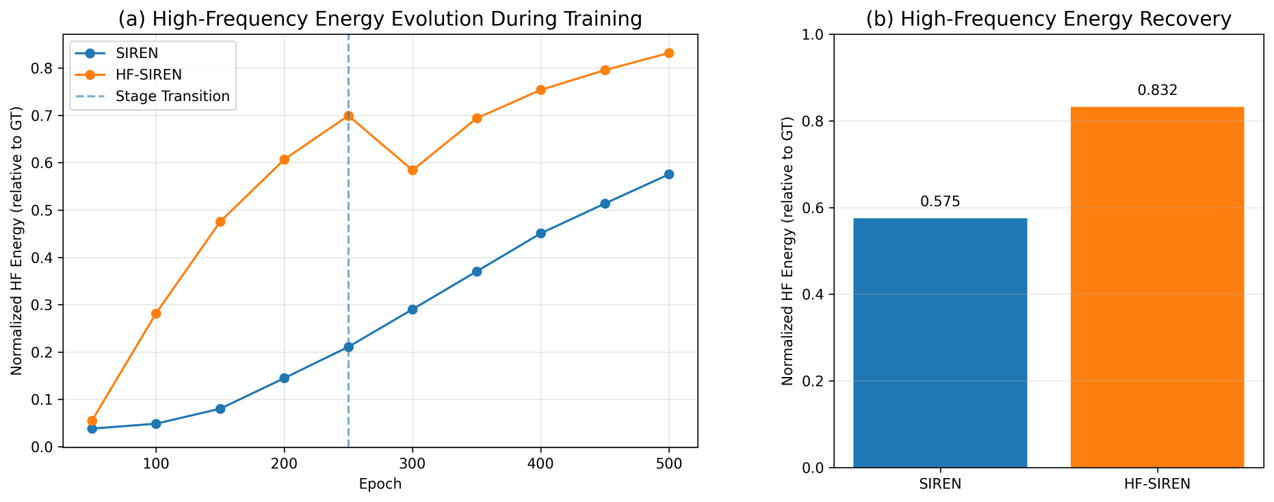}
\caption{High-frequency energy analysis on \texttt{kodim14}. (a) Normalized high-frequency (HF) energy versus epoch for SIREN and HF-SIREN, where the dashed line denotes the transition from Stage 1 to Stage 2. (b) Comparison of final HF energy recovery relative to the ground truth.}
\label{fig:FFTAnalysis2}
\vspace{-3.5mm}
\end{figure}

\textbf{High-Frequency Energy Analysis.} We next quantify the impact of HF-INR on high-frequency (HF) components for \texttt{kodim14} by analyzing the evolution of normalized HF energy during training and its final recovery. Fig.~\ref{fig:FFTAnalysis2}(a) illustrates the normalized HF energy curve during training, where the HF energy is computed by aggregating the Fourier magnitudes beyond 50\% of the maximum radial distance from the spectrum center. During Stage 1, the HF-SIREN curve exhibits a substantially steeper rise than the baseline, reflecting the effect of the proposed high-frequency-first optimization strategy. Following the transition to full-image optimization in Stage 2, a temporary drop is observed due to the shift in optimization focus, after which the curve steadily rises as the model progressively refines both low- and high-frequency components. Fig.~\ref{fig:FFTAnalysis2}(b) presents the final normalized HF energy recovery relative to the ground truth. HF-SIREN achieves a recovery of $0.832$, substantially higher than the $0.575$ achieved by SIREN. Overall, these results demonstrate that HF-INR not only accelerates early high-frequency learning but also achieves superior final high-frequency energy recovery, leading to more faithful reconstruction of fine details.
\vspace{-1mm}

\begin{table}[t]
\centering
\caption{Ablation results for Stage~1 training epochs of HF-SIREN on the \texttt{Kodak} dataset. Best and second-best results are highlighted in bold and underlined, respectively.}
\label{tab:stage1_epochs}
\resizebox{.825\linewidth}{!}{%
\begin{tabular}{c|c|c|c|c|c}
\toprule
Epochs & 150 & 200 & 250 & 300 & 350 \\
\midrule
PSNR & 36.46 & 36.55 & \textbf{36.78} & \underline{36.67} & 36.31 \\
SSIM & 0.9489 & 0.9495 & \textbf{0.9517} & \underline{0.9499} & 0.9462 \\
\bottomrule
\end{tabular}
}
\vspace{-2.5mm}
\end{table}

%First, we investigate the impact of the Stage 1 training duration for HF-SIREN, which controls how long the model is optimized under HF-prioritized supervision.
\subsection{Ablation Studies}
\textbf{Effect of Stage 1 Epochs.} First, we investigate the impact of the Stage 1 training duration for HF-SIREN on the \texttt{Kodak} dataset. As shown in Table~\ref{tab:stage1_epochs}, increasing the number of Stage 1 epochs leads to consistent performance gains up to an optimal point, with PSNR improving from $36.46$ dB at 150 epochs to $36.78$ dB at 250 epochs, alongside a corresponding rise in SSIM from $0.9489$ to $0.9517$. This trend indicates that allocating sufficient iterations to Stage 1 enables the INR to better capture sharp edges and fine local variations before transitioning to full-image optimization. However, further increasing the epochs results in a gradual performance decline (e.g., PSNR drops to $36.31$ dB at 350 epochs), suggesting that overly prolonged Stage 1 training biases the model toward masked regions and weakens its ability to balance high- and low-frequency components during Stage 2. Considering this trade-off, we set the Stage 1 duration to 250 epochs to achieve an effective balance between early HF emphasis and subsequent global refinement.

\begin{table}[t]
\centering
\caption{Ablation results for different combinations of $\tau$ and $n$ for HF-SIREN on the \texttt{Kodak} dataset. Best and second-best results are highlighted in bold and underlined, respectively.}
\label{tab:ablation_tau_n}
\resizebox{.85\linewidth}{!}{%
\begin{tabular}{c|cc|cc|cc}
\toprule
\multirow{2}{*}{$\tau$} & \multicolumn{2}{c|}{$n = 4$} & \multicolumn{2}{c|}{$n = 8$} & \multicolumn{2}{c}{$n = 12$} \\
 & PSNR & SSIM & PSNR & SSIM & PSNR & SSIM \\
\midrule
0.3 & 36.21 & 0.9477 & 36.22 & 0.9486 & 34.74 & 0.9366 \\
0.4 & 36.48 & 0.9492 & \underline{36.69} & \underline{0.9509} & 34.98 & 0.9381 \\
0.5 & 36.50 & 0.9493 & \textbf{36.78} & \textbf{0.9517} & 35.61 & 0.9451 \\
0.6 & 35.97 & 0.9449 & 36.65 & 0.9495 & 35.57 & 0.9440 \\
0.7 & 35.46 & 0.9387 & 36.16 & 0.9453 & 35.56 & 0.9424 \\
\bottomrule
\end{tabular}
}
\vspace{-3.25mm}
\end{table}

\textbf{Effect of $\tau$ and $n$.}
We also examine the sensitivity of our proposed training strategy to the high-frequency threshold $\tau$ and neighborhood size $n$ using HF-SIREN on the \texttt{Kodak} dataset. As shown in Table~\ref{tab:ablation_tau_n}, performance varies only slightly over a moderate range of $\tau$, with the best results obtained around $\tau = 0.5$ for different neighborhood sizes. This behavior indicates that the soft masking mechanism is not strictly dependent on the exact threshold choice, as long as $\tau$ separates strong local variations from smoother regions. In contrast, the neighborhood size has a more noticeable effect. A local context of $n = 8$ consistently yields the strongest performance (e.g., $36.78$ dB PSNR), whereas using an excessively large neighborhood ($n = 12$) leads to a clear degradation. This suggests that overly large neighborhoods may dilute fine-grained variations by incorporating less relevant pixels into the soft mask computation. Consequently, we adopt $n = 8$ and $\tau = 0.5$ as the default configuration for our experiments, as it provides a balanced trade-off between local detail preservation and global reconstruction accuracy.

\textbf{Soft Mask vs. Hard Selection.} To further investigate the role of our proposed neighbor-aware soft mask, we conduct an ablation study by replacing it with a hard high-frequency selection strategy during Stage~1. Specifically, pixels whose maximum neighbor-wise difference $\Delta(i,j)$ exceeds the threshold $\tau = 0.5$ are selected, while the remaining pixels are ignored. Formally, the hard mask is defined as:
\begin{equation}
\mathcal{M}_{\text{HF}}^{\text{hard}}(i,j)=
\begin{cases}
1, & \text{if } \Delta(i,j) > \tau,\\
0, & \text{otherwise}.
\end{cases}
\end{equation}
%During Stage~1, the same loss as in Eq.~\ref{eq:hf_loss} is used, with $\mathcal{M}_{\text{HF}}$ replaced by $\mathcal{M}_{\text{HF}}^{\text{hard}}$. Stage~2 training remains unchanged, using the standard mean squared error (MSE) loss over all pixels. Table~\ref{tab:kodak_soft_vs_hard} reports the ablation results on the KODAK dataset. Restricting Stage~1 to hard-selected pixels consistently improves reconstruction quality over the corresponding vanilla INRs across all three backbones, confirming that early prioritization of high-frequency regions is already beneficial.
During Stage~1, the same loss as in Eq.~\ref{eq:hf_loss} is used, with $\mathcal{M}_{\text{HF}}$ replaced by $\mathcal{M}_{\text{HF}}^{\text{hard}}$. All other settings remain unchanged. Table~\ref{tab:kodak_soft_vs_hard} reports the ablation results on the \texttt{Kodak} dataset. Restricting Stage~1 to hard-selected pixels consistently improves reconstruction quality over the corresponding vanilla INRs, confirming that early prioritization of high-frequency regions is already beneficial. However, the proposed soft mask consistently yields higher PSNR and SSIM in all cases. For instance, HF-SIREN improves PSNR from $33.51$ dB to $34.86$ dB with hard selection, while using the soft mask further boosts performance to $36.78$ dB. Similar trends are also observed for FINER and SCONE, with the soft-mask approach giving noticeably larger gains. Overall, these results show that while hard selection already helps mitigate spectral bias, the soft mask provides more stable optimization during the early stage and leads to superior reconstruction quality.

\begin{table}[t]
\centering
\caption{Ablation results on the \texttt{Kodak} dataset for evaluating the effect of the soft mask. Green subscripts indicate improvements over the corresponding vanilla backbone, while red superscripts denote performance drops relative to the soft-mask variant. Best results are highlighted in bold.}
\label{tab:kodak_soft_vs_hard}
\resizebox{.80\linewidth}{!}{%
\begin{tabular}{c|cc}
\toprule
\multirow{2}{*}{Model} & \multicolumn{2}{c}{KODAK}\\
 & PSNR & SSIM \\
\midrule
SIREN & 33.51 & 0.9185 \\

$\text{HF-SIREN}_{\text{(soft mask)}}$ &
$\textbf{36.78}_{\impr{\text{\scriptsize(+3.27)}}}$ &
$\textbf{0.9517}_{\impr{\text{\scriptsize(+0.0332)}}}$ \\

$\text{HF-SIREN}_{\text{(w/o soft mask)}}$ &
$\text{34.86}^{\textcolor{red}{\text{\scriptsize(-1.92)}}}_{\impr{\text{\scriptsize(+1.35)}}}$ &
$\text{0.9332}^{\textcolor{red}{\text{\scriptsize(-0.0185)}}}_{\impr{\text{\scriptsize(+0.0147)}}}$ \\
\midrule

FINER & 37.07 & 0.9589 \\

$\text{HF-FINER}_{\text{(soft mask)}}$ &
$\textbf{40.22}_{\impr{\text{\scriptsize(+3.15)}}}$ &
$\textbf{0.9746}_{\impr{\text{\scriptsize(+0.0157)}}}$ \\

$\text{HF-FINER}_{\text{(w/o soft mask)}}$ &
$\text{38.24}^{\textcolor{red}{\text{\scriptsize(-1.98)}}}_{\impr{\text{\scriptsize(+1.17)}}}$ &
$\text{0.9634}^{\textcolor{red}{\text{\scriptsize(-0.0112)}}}_{\impr{\text{\scriptsize(+0.0045)}}}$ \\
\midrule

SCONE & 33.18 & 0.9403 \\

$\text{HF-SCONE}_{\text{(soft mask)}}$ &
$\textbf{40.66}_{\impr{\text{\scriptsize(+7.48)}}}$ &
$\textbf{0.9795}_{\impr{\text{\scriptsize(+0.0392)}}}$ \\

$\text{HF-SCONE}_{\text{(w/o soft mask)}}$ &
$\text{36.29}^{\textcolor{red}{\text{\scriptsize(-4.37)}}}_{\impr{\text{\scriptsize(+3.11)}}}$ &
$\text{0.9547}^{\textcolor{red}{\text{\scriptsize(-0.0248)}}}_{\impr{\text{\scriptsize(+0.0144)}}}$ \\
\bottomrule
\end{tabular}
}
\vspace{-3.25mm}
\end{table}

\textbf{Comparison with Alternative Training Strategies.} To further investigate whether the observed improvements are specific to the proposed neighbor-aware soft mask, we compare our method with several alternative training strategies. Specifically, we consider Prediction-aware, Saliency-based, and FFT-based INRs in both one-stage and two-stage variants, as well as a one-stage variant of our proposed HF-INR.
%To further investigate whether the observed improvements are specific to the proposed neighbor-aware soft mask, we compare our method with several alternative training strategies that also emphasize visually important regions. Specifically, we consider edge-aware and saliency-based INRs in both one-stage and two-stage variants, as well as a one-stage variant of our proposed HF-INR.
%To further verify whether the proposed improvement comes from the specific neighbor-aware HF prioritization rather than simply from adding extra structural guidance, we compare our method with alternative training strategies that also emphasize visually important regions. In particular, we consider edge-aware and saliency-based INR variants in both one-stage and two-stage forms, as well as a one-stage version of our HF-based weighting.

\textit{Prediction-aware INR.} We first consider a prediction-aware strategy inspired by focal loss, where pixels with larger reconstruction errors receive greater optimization emphasis. Specifically, the prediction-aware weight is computed as:
%We first consider a Prediction-aware INR inspired by focal loss, where pixels with larger reconstruction errors receive greater optimization emphasis. Specifically, the prediction-aware weight is defined as:
% \begin{equation}
% w_{\mathrm{pred}}(i,j)=\left|\psi_{\Theta}(i,j)-I(i,j)\right|^{\gamma},
% \end{equation}
\begin{equation}
\widetilde{w}_{\mathrm{pred}}(i,j) = \left|\psi_{\Theta}(i,j)-I(i,j)\right|^{\gamma},
\end{equation}
where $\gamma$ controls the degree of emphasis on difficult pixels. The weights are normalized to preserve a comparable loss scale, and the weighted reconstruction objective is given by:
\begin{equation}
\mathcal{L}_{\mathrm{pred}}=\mathbb{E}_{i,j}\left[w_{\mathrm{pred}}(i,j)\left\|\psi_{\Theta}(i,j)-I(i,j)\right\|_2^2\right].
\end{equation}

\textit{Saliency-based INR.} We next consider a saliency-based strategy, where pixels with higher local contrast receive greater optimization emphasis. Specifically, the center-surround contrast is computed as:
\begin{equation}
M_{\mathrm{con}}
=
\sqrt{\frac{1}{C}\sum_{c=1}^{C}
\left(I_c-\bar{I}_c\right)^2},
\end{equation}
%We next consider a Saliency-based INR, where pixels with stronger local visual contrast receive greater optimization emphasis. Specifically, the saliency map is computed using center-surround contrast as:
% \begin{equation}
% M_{\mathrm{sal}}=\sqrt{\frac{1}{C}\sum_{c=1}^{C}\left(I_c-\bar{I}_c\right)^2},
% \end{equation}
where $\bar{I}$ denotes the local mean image obtained through average pooling. The contrast map is then spatially aggregated and normalized to form $M_{\mathrm{sal}}$, followed by the construction of dense saliency weights as:
\begin{equation}
\widetilde{w}_{\mathrm{sal}}(i,j) = 1+M_{\mathrm{sal}}(i,j),
\end{equation}
which are normalized to preserve a comparable loss scale. The weighted reconstruction objective is given by:
\begin{equation}
\mathcal{L}_{\mathrm{sal}} = \mathbb{E}_{i,j} \left[ w_{\mathrm{sal}}(i,j) \left\| \psi_{\Theta}(i,j)-I(i,j) \right\|_2^2 \right].
\end{equation}
%where $\bar{I}$ denotes the local mean image obtained through average pooling. The contrast map is spatially aggregated and normalized to obtain $M_{\mathrm{sal}}$. Dense saliency weights are subsequently constructed as
%where $\bar{I}$ denotes the local mean image obtained using average pooling. The saliency map is further spatially aggregated and normalized to obtain the prediction weights. The weighted reconstruction objective is then defined as:

\textit{FFT-based INR.} We also consider an FFT-based strategy, where pixels with stronger high-frequency responses receive greater optimization emphasis. Specifically, the FFT-based importance map is computed as:
\begin{equation}
\widetilde{M}_{\mathrm{fft}} = \frac{1}{C} \sum_{c=1}^{C} \left|  \mathcal{F}^{-1} \!\left(  R\odot\mathcal{F}(I_c)  \right) \right|,
\end{equation}
% \begin{equation}
% M_{\mathrm{fft}}=
% \frac{1}{C}
% \sum_{c=1}^{C}
% \left|
% \mathcal{F}^{-1}
% \!\left(
% R\odot\mathcal{F}(I_c)
% \right)
% \right|,
% \end{equation}
where $\mathcal{F}$ and $\mathcal{F}^{-1}$ denote the Fourier and inverse Fourier transforms, respectively, and $R$ is a radial high-pass weighting mask. The importance map is then normalized to form $M_{\mathrm{fft}}$, followed by the construction of dense FFT-based weights as:
\begin{equation}
\widetilde{w}_{\mathrm{fft}}(i,j)=1+M_{\mathrm{fft}}(i,j),
\end{equation}
which are normalized to preserve a comparable loss scale. The weighted reconstruction objective is given by:
\begin{equation}
\mathcal{L}_{\mathrm{fft}} =  \mathbb{E}_{i,j} \left[  w_{\mathrm{fft}}(i,j) \left\|  \psi_{\Theta}(i,j)-I(i,j) \right\|_2^2  \right].
\end{equation}

\begin{table*}[t]
\centering
\caption{Comparison of different training strategies for SIREN, FINER, and SCONE on the \texttt{Kodak} dataset. Best and second-best results for each backbone are highlighted in bold and underlined, respectively.}
\label{tab:alternative_training}

\resizebox{\textwidth}{!}{%
\begin{tabular}{c|cc|cccc|cccc|cccc|cccc}
\toprule

\multirow{3}{*}{Model}

& \multicolumn{2}{c|}{Standard}

& \multicolumn{4}{c|}{Prediction-aware}

& \multicolumn{4}{c|}{Saliency-based}

& \multicolumn{4}{c|}{FFT-based}

& \multicolumn{4}{c}{HF-prioritized (Ours)} \\

&
\multicolumn{2}{c|}{Training}
& \multicolumn{2}{c}{1-stage}
& \multicolumn{2}{c|}{2-stage}

& \multicolumn{2}{c}{1-stage}
& \multicolumn{2}{c|}{2-stage}

& \multicolumn{2}{c}{1-stage}
& \multicolumn{2}{c|}{2-stage}

& \multicolumn{2}{c}{1-stage}
& \multicolumn{2}{c}{2-stage} \\

\cmidrule(lr){2-3}
\cmidrule(lr){4-5}
\cmidrule(lr){6-7}
\cmidrule(lr){8-9}
\cmidrule(lr){10-11}
\cmidrule(lr){12-13}
\cmidrule(lr){14-15}
\cmidrule(lr){16-17}
\cmidrule(lr){18-19}

& PSNR & SSIM
& PSNR & SSIM
& PSNR & SSIM
& PSNR & SSIM
& PSNR & SSIM
& PSNR & SSIM
& PSNR & SSIM
& PSNR & SSIM
& PSNR & SSIM \\

\midrule

SIREN
& 33.51 & 0.9185
& 33.71 & 0.9073
& 34.38 & 0.9231
& 34.84 & 0.9284
& 35.02 & 0.9326
& 35.20 & 0.9346
& \underline{35.58} & \underline{0.9387}
& 33.98 & 0.8957
& \textbf{36.78} & \textbf{0.9517} \\

FINER
& 37.07 & 0.9589
& 37.55 & 0.9555
& 37.85 & 0.9598
& 38.44 & 0.9622
& 38.71 & 0.9653
& 38.79 & 0.9663
& \underline{39.18} & \underline{0.9686}
& 36.19 & 0.9248
& \textbf{40.22} & \textbf{0.9746} \\

SCONE
& 33.18 & 0.9403
& 34.57 & 0.9411
& \underline{34.64} & 0.9465
& 34.15 & 0.9365
& 34.43 & 0.9475
& 34.01 & 0.9392
& 34.46 & \underline{0.9493}
& 28.17 & 0.8407
& \textbf{40.66} & \textbf{0.9795} \\

\bottomrule
\end{tabular}
}
\vspace{-2mm}
\end{table*}

\begin{table*}[t]
\centering
\caption{Seed-wise comparison across SIREN, FINER, and SCONE on the \texttt{Kodak} dataset.}
\label{tab:rebuttal_seed}
\resizebox{.8\linewidth}{!}{%
\begin{tabular}{c | cc | cc | cc | cc | cc | cc}
\toprule
\multirow{2}{*}{Model}
& \multicolumn{2}{c|}{Seed = 1}
& \multicolumn{2}{c|}{Seed = 29}
& \multicolumn{2}{c|}{Seed = 145}
& \multicolumn{2}{c|}{Seed = 1064}
& \multicolumn{2}{c|}{Seed = 13778}
& \multicolumn{2}{c}{mean $\pm$ std} \\
\cmidrule(lr){2-3}
\cmidrule(lr){4-5}
\cmidrule(lr){6-7}
\cmidrule(lr){8-9}
\cmidrule(lr){10-11}
\cmidrule(lr){12-13}
& PSNR & SSIM
& PSNR & SSIM
& PSNR & SSIM
& PSNR & SSIM
& PSNR & SSIM
& PSNR & SSIM \\
\midrule

SIREN
& 33.57 & 0.9197
& 33.38 & 0.9177
& 33.37 & 0.9173
& 33.44 & 0.9182
& 33.53 & 0.9206
& 33.46 $\pm$ 0.09
& 0.9187 $\pm$ 0.0014 \\

HF-SIREN
& 36.85 & 0.9529
& 36.63 & 0.9496
& 36.66 & 0.9503
& 36.64 & 0.9503
& 36.61 & 0.9503
& 36.68 $\pm$ 0.10
& 0.9507 $\pm$ 0.0013 \\

\midrule

FINER
& 37.04 & 0.9588
& 37.15 & 0.9592
& 37.07 & 0.9593
& 37.01 & 0.9584
& 37.26 & 0.9599
& 37.11 $\pm$ 0.10
& 0.9591 $\pm$ 0.0006 \\

HF-FINER
& 40.44 & 0.9756
& 40.42 & 0.9750
& 40.33 & 0.9749
& 40.05 & 0.9739
& 40.19 & 0.9740
& 40.29 $\pm$ 0.16
& 0.9747 $\pm$ 0.0007 \\

\midrule

SCONE
& 33.26 & 0.9408
& 33.32 & 0.9412
& 33.05 & 0.9387
& 33.08 & 0.9397
& 33.33 & 0.9411
& 33.21 $\pm$ 0.13
& 0.9403 $\pm$ 0.0011 \\

HF-SCONE
& 40.64 & 0.9795
& 40.80 & 0.9801
& 40.52 & 0.9790
& 40.57 & 0.9790
& 40.68 & 0.9794
& 40.64 $\pm$ 0.11
& 0.9794 $\pm$ 0.0005 \\

\bottomrule
\end{tabular}
}
\vspace{-2.5mm}
\end{table*}

\textit{One-stage HF-INR.} Finally, we consider a one-stage variant of the proposed HF-INR to compare with the proposed two-stage optimization strategy. In this setting, the same neighbor-aware soft mask defined in Eq.~(10) is applied for the entire 500-epoch training, without transitioning to standard full-image optimization. Therefore, pixels associated with stronger local intensity variations continue to receive higher optimization emphasis, while the remaining pixels are still optimized with lower weights.
%\textit{One-stage HF-INR.} Finally, we consider a one-stage variant of the proposed HF-INR to compare with the proposed two-stage optimization strategy. In this setting, the same neighbor-aware soft mask defined in Eq.~(10) is applied for the entire 500-epoch training, without transitioning to standard full-image optimization. All other parameters remain identical to those of the proposed two-stage approach. Consequently, pixels associated with stronger local intensity variations continue to receive higher optimization emphasis, while the remaining pixels are still optimized with lower weights. This comparison allows us to evaluate the effect of maintaining high-frequency prioritization throughout the full 500-epoch training, as opposed to restricting it to the initial 250 epochs before switching to full-image optimization.\\
%\textit{One-stage HF-INR.} Finally, we also consider a one-stage variant of the proposed HF-INR to isolate the effect of the two-stage training strategy. In this setting, the same neighbor-aware soft mask described in Eq.~(10) is applied throughout the entire training process, without any stage transition. That is, the model is continuously guided to prioritize high-frequency regions using the soft weighting mechanism for all training epochs. This variant allows us to examine whether the performance gain arises solely from pixel-wise high-frequency weighting, or from the staged optimization strategy that progressively shifts the learning focus from high-frequency regions to full-image reconstruction.\\

\textit{Experimental Setup.} We evaluate the alternative training strategies on the \texttt{Kodak} dataset. During training, the one-stage variants optimize their corresponding reconstruction objectives throughout the entire 500-epoch schedule, whereas the two-stage variants optimize the same objectives over the first 450 epochs (Stage~1), followed by standard MSE optimization for the remaining 50 epochs (Stage~2). Unless otherwise specified, all other settings remain identical to those described in Section~\ref{sec:exp}. Additional implementation details, algorithms, and empirical analyses supporting the selected Stage~1 duration are provided in the supplementary material.

\textit{Performance Comparison.} Table~\ref{tab:alternative_training} compares different training strategies across SIREN, FINER, and SCONE. Among the alternatives, the FFT-based strategy achieves the best performance for SIREN ($35.58$ dB/$0.9387$ SSIM) and FINER ($39.18$ dB/$0.9686$ SSIM), whereas for SCONE, the Prediction-aware strategy achieves the highest PSNR ($34.64$ dB) while the FFT-based strategy achieves the highest SSIM ($0.9493$). This variation indicates that the effectiveness of different strategies varies across INR backbones and evaluation metrics. The one-stage variants, on the other hand, consistently yield lower performance than their corresponding two-stage counterparts. In contrast, the proposed HF-INR achieves the best performance across all three backbones, outperforming the strongest alternative strategy by $1.20$ dB, $1.04$ dB, and $6.02$ dB on SIREN, FINER, and SCONE, respectively, while also obtaining the highest SSIM in every case.
\begin{figure*}[t]
\centering
\includegraphics[width=.92\linewidth]{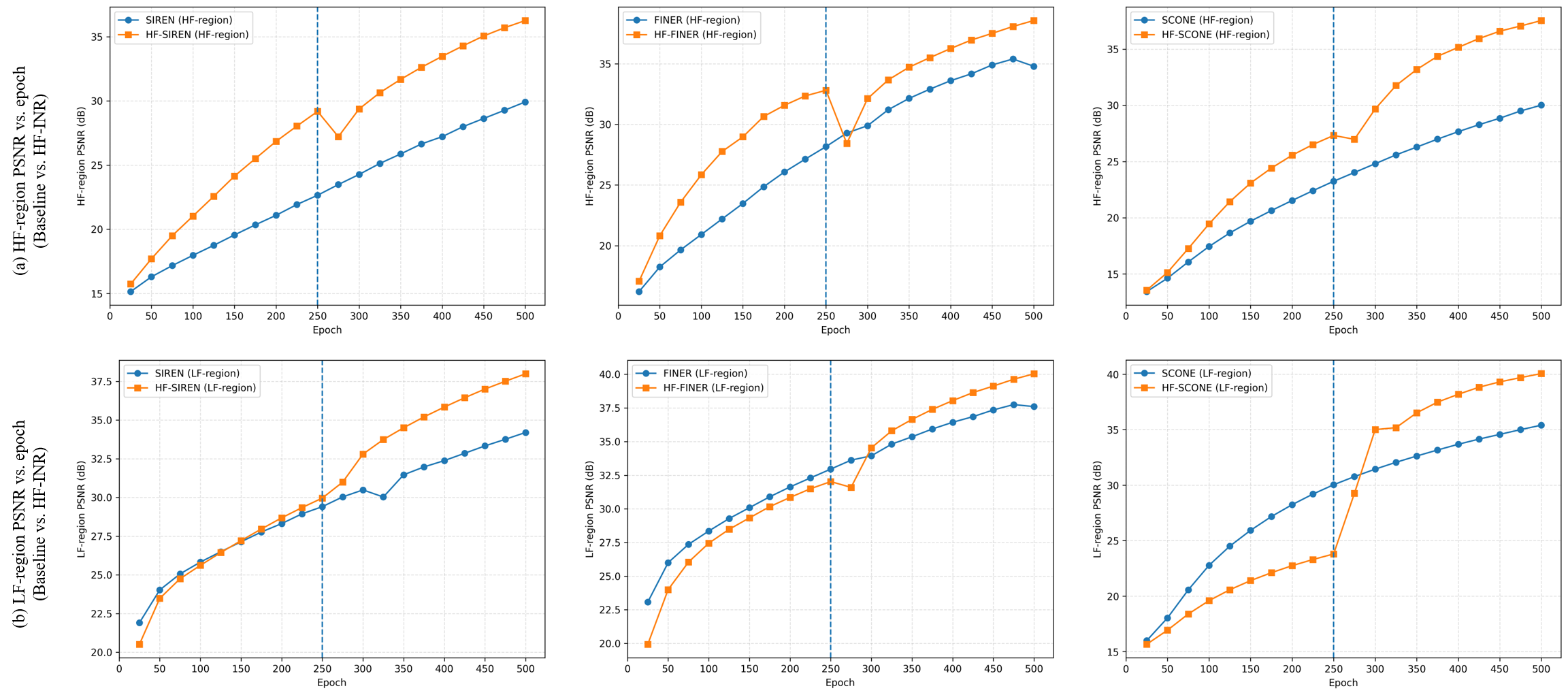}
\caption{Region-wise PSNR trajectories over training epochs on \texttt{kodim21}. (a) HF-region PSNR versus epoch for baseline INRs and their HF-INR counterparts. (b) Corresponding LF-region PSNR versus epoch. The vertical dashed line denotes the transition from HF-prioritized training (Stage 1) to standard full-image optimization (Stage 2) at epoch 250.}
\label{fig:finalq}
\vspace{-4.35mm}
\end{figure*}
%\textbf{Robustness Analysis Across Random Seeds.} To verify that the observed performance gains are not due to randomness, we evaluate the baseline and the proposed method across multiple random seeds. As shown in Table~\ref{tab:rebuttal_seed}, the results are highly consistent for all three backbones. For instance, SIREN achieves $33.51 \pm 0.04$ PSNR, while the proposed variant reaches $36.78 \pm 0.04$, indicating a stable and significant improvement. Similar trends are observed for FINER ($37.07 \pm 0.04$ vs. $40.22 \pm 0.04$) and SCONE ($33.21 \pm 0.07$ vs. $40.63 \pm 0.05$). Notably, the standard deviations remain very small across all cases, confirming that the improvements are consistent across different initializations and are not driven by random variations. These results demonstrate the robustness and reliability of the proposed method.
%To verify that the performance gains are consistent across different initializations, we evaluate the baseline models and their HF variants across five independent random seeds.

\textbf{Robustness Analysis Across Random Seeds.} To verify that the performance gains are consistent across different initializations, we evaluate the baseline models and their HF variants on the \texttt{Kodak} dataset across five independent random seeds. As summarized in Table~\ref{tab:rebuttal_seed}, the proposed strategy improves all three INR backbones with low variance across different seeds. For instance, the baseline SIREN achieves $33.46 \pm 0.09$ dB PSNR and $0.9187 \pm 0.0014$ SSIM, while HF-SIREN achieves $36.68 \pm 0.10$ dB PSNR and $0.9507 \pm 0.0013$ SSIM. Similarly, FINER improves from $37.11 \pm 0.10$ dB / $0.9591 \pm 0.0006$ to $40.29 \pm 0.16$ dB / $0.9747 \pm 0.0007$, and SCONE improves from $33.21 \pm 0.13$ dB / $0.9403 \pm 0.0011$ to $40.64 \pm 0.11$ dB / $0.9794 \pm 0.0005$. Furthermore, the mean PSNR across the five random seeds is consistent with the representative results reported in Table~\ref{tab:kodak_Urban100_flipped}, where HF-SIREN, HF-FINER, and HF-SCONE achieve $36.78$, $40.22$, and $40.66$ dB PSNR, respectively. Similar consistency is also observed for SSIM. Notably, the standard deviations remain low for both PSNR and SSIM across all three backbones, confirming that the improvements are reproducible rather than being artifacts of a favorable seed.
\begin{table}[t]
\centering\caption{Region-wise PSNR comparison on \texttt{kodim21}. Best results are highlighted in bold. $\Delta$ indicates the performance gain of HF variants over their vanilla counterparts, reported as +HF PSNR (dB) / +LF PSNR (dB).}
\label{tab:kodim21_region}
\resizebox{.7175\linewidth}{!}{%
\begin{tabular}{c|cc|c}
\toprule
\multirow{2}{*}{Model} & \multicolumn{2}{c|}{kodim21} & \multirow{2}{*}{$\Delta$} \\
& HF PSNR & LF PSNR & \\
\midrule
SIREN     & 29.92 & 34.20 & -- \\
\textbf{HF-SIREN}  & \textbf{36.26} & \textbf{37.99} & \impr{+6.34 / +3.79} \\
\midrule
FINER     & 34.79 & 37.60 & -- \\
\textbf{HF-FINER}  & \textbf{38.56} & \textbf{40.04} & \impr{+3.77 / +2.44} \\
\midrule
SCONE     & 30.02 & 35.40 & -- \\
\textbf{HF-SCONE}  & \textbf{37.55} & \textbf{40.05} & \impr{+7.53 / +4.65} \\
\bottomrule
\end{tabular}
}
\vspace{-4mm}
\end{table}

\textbf{Why Does HF-SCONE Benefit More?} The proposed HF-INR framework consistently improves the performance of all three INR backbones. However, compared to HF-SIREN and HF-FINER, HF-SCONE exhibits substantially larger performance gains across the evaluated datasets. For example, on the KODAK dataset, HF-SCONE improves its corresponding baseline PSNR by $+7.48$ dB, whereas HF-SIREN and HF-FINER achieve gains of $+3.27$ dB and $+3.15$ dB, respectively. A plausible explanation can be derived from the optimization characteristics of the SCONE architecture. Unlike conventional coordinate-based INRs that directly optimize a globally shared implicit representation, SCONE jointly learns geometry-dependent spatial masks and localized Random Fourier Feature (RFF) representations. Specifically, at each layer, the spatial mask is generated as:
\begin{equation}
\mathcal{M}_{\ell}=\sigma\!\left(W_{\ell-1}\mathbf{z}_{\ell-1}+\mathbf{b}_{\ell-1}\right),
\end{equation}
which is then applied to the corresponding Fourier basis as:
\begin{equation}
\mathbf{z}_{\ell}=\mathcal{M}_{\ell}\odot g_{\ell}(\mathbf{x};\omega_{\ell}),
\end{equation}
where $g_{\ell}(\mathbf{x};\omega_{\ell})$ denotes the layer-wise RFF representation and $\odot$ represents element-wise multiplication. The spatial masks $\mathcal{M}_{\ell}$ determine where each Fourier basis is activated, making the localized Fourier representations directly dependent on these masks. Unlike SCONE, HF-INR neither modifies the network architecture nor introduces additional learnable components. Instead, it changes the optimization trajectory by emphasizing high-frequency regions during the initial stage before transitioning to full-image optimization. We conjecture that this early emphasis provides a more informative supervisory signal while the spatial masks are still being learned, enabling improved joint optimization of the masks and localized Fourier representations. This allows the subsequent full-image optimization stage to better exploit SCONE's spatially adaptive representation capability, leading to larger performance gains than those observed for HF-SIREN and HF-FINER.

\begin{table}[t]
\centering
\caption{Average training and inference time per image on the \texttt{Kodak} dataset. $\Delta$ reports the additional training and inference overhead of HF variants compared to their vanilla counterparts.}
\label{tab:time}
\resizebox{0.823\linewidth}{!}{%
\begin{tabular}{c|cc|c}
\toprule
\multirow[c]{3}{*}{Model} & \multicolumn{2}{c|}{KODAK} & \multirow[c]{3}{*}{$\Delta$} \\
 & Training Time & Inference Time &  \\
 & (s / image) & (s / image) &  \\
\midrule
SIREN      & 11.8031 & 0.0034 & -- \\
HF-SIREN   & 11.8732 & 0.0034 & +0.0701 / +0 \\
\midrule
FINER      & 13.4128 & 0.0062 & -- \\
HF-FINER   & 13.4163 & 0.0062 & +0.0035 / +0 \\
\midrule
SCONE      & 18.0530 & 0.0080 & -- \\
HF-SCONE   & 18.1837 & 0.0080 & +0.1307 / +0 \\
\bottomrule
\end{tabular}
}
\vspace{-4.5mm}
\end{table}

\textbf{A Second Look on the Performance Gain.}
In the earlier evaluations, we analyzed INR performance on entire images and showed that the proposed strategy yields substantial gains. To better understand how these gains emerge during training, we examine region-wise PSNR trajectories over epochs on \texttt{kodim21}. Fig.~\ref{fig:finalq}(a) shows the HF-region PSNR vs. epoch curves for the baselines and their HF variants, while Fig.~\ref{fig:finalq}(b) reports the corresponding LF-region PSNR vs. epoch curves. HF and LF regions are defined using a neighbor-based soft mask $\mathcal{M}_{\text{HF}}$ derived from the ground-truth image using the same criterion as in Stage~1 and kept fixed across all models. As shown in Fig.~\ref{fig:finalq}(a), emphasizing HF components in the initial stage consistently accelerates HF-region reconstruction across all three backbones, with HF-INR variants achieving substantially higher HF-region PSNR in early epochs. After the transition to full-image optimization at epoch 250, a temporary drop in HF-region PSNR is observed, followed by continued improvement. This behavior suggests that prioritizing HF regions early alters the optimization dynamics, enabling fine details to be captured before the model adapts to full-image reconstruction. Meanwhile, Fig.~\ref{fig:finalq}(b) indicates that LF-region reconstruction is not compromised. Although LF PSNR starts slowly during the early stage, it improves steadily over training and eventually surpasses the baselines for all backbones. To quantify the final outcome, we report region-wise PSNR at convergence in Table~\ref{tab:kodim21_region}. HF-INR variants consistently outperform their vanilla counterparts in both HF and LF regions. For example, HF-SIREN boosts HF-region PSNR from $29.92$ dB to $36.26$ dB, while LF-region PSNR increases from $34.20$ dB to $37.99$ dB, with similar improvements observed for FINER and SCONE.

\textbf{Runtime Analysis.} Table~\ref{tab:time} reports the average training and inference time per image on the KODAK dataset. HF-enhanced models introduce only a marginal increase in training time (less than 1\% across all backbones) due to the neighbor-aware soft-mask computation in Stage~1. Inference time remains unchanged for all methods, as the proposed strategy does not modify the network architecture and mainly affects the training dynamics.
\vspace{-2.5mm}
\section{Conclusion}
\label{sec:concl}
Recognizing that most INR papers focus on modifying internal components such as network architectures or activation functions, our method takes a complementary step toward improving performance through training dynamics. We propose a two-stage training strategy that leverages a neighbor-aware soft mask to adaptively emphasize high-frequency regions in the early training phase. This simple yet effective strategy helps mitigate spectral bias by encouraging the model to capture fine details before generalizing to the entire image. Experimental results on natural images and CT scans using multiple INR backbones demonstrate consistent improvements in both PSNR and SSIM. Our investigation reveals that the when and where of learning can be just as critical as the design of the model itself. While this work focuses on image implicit neural representations, extending the proposed training strategy to other modalities is left for future work.
\vspace{-1.5mm}

\bibliography{HFINR}
\bibliographystyle{IEEEtran}

\vfill

\end{document}